\documentclass[a4paper, 10pt, conference]{ieeeconf}      
          
\usepackage{FG2021}
\usepackage{multirow}
\usepackage{makecell}
\usepackage[pdftex]{graphicx}
\usepackage{amsmath}
\usepackage{algorithmic}
\usepackage{cite}
\usepackage[linesnumbered,ruled,vlined]{algorithm2e}
\usepackage{arydshln}
\usepackage{hyperref}
 \hypersetup{
     colorlinks=true,
     linkcolor=blue,
     }
\usepackage[accsupp]{axessibility} 

\FGfinalcopy 

\IEEEoverridecommandlockouts                              
\def\FGPaperID{287} 

\title{Similarity-based Gray-box Adversarial Attack Against Deep Face Recognition}

\author{\parbox{16cm}{\centering
    {\large Hanrui Wang$^1$, Shuo Wang$^2$, Zhe Jin$^1$, Yandan Wang$^3$, Cunjian Chen$^{4,5}$, Massimo Tistarelli$^6$}\\
    {\normalsize
    $^1$ School of Information Technology, Monash University Malaysia, Malaysia\\
    $^2$ Data61, CSIRO, Australia\\
    $^3$ Faculty of Computer Science and Technology, Wenzhou University, China\\
    $^4$ Monash Suzhou Research Institute, China\\
    $^5$ Department of Data Science and AI, Monash University, Australia\\
    $^6$ The University of Sassari, Italy}}
    \thanks{This work was supported by grants from Ministry of Higher Education (MOHE) Malaysia through Fundamental Research Grant Scheme (FRGS/1/2018/ICT02/MUSM/03/3).}
}

\begin{document}
%
%
%




\IEEEoverridecommandlockouts\pubid{\makebox[\columnwidth]{978-1-6654-3176-7/21/\$31.00~\copyright{}2021 IEEE \hfill}
\hspace{\columnsep}\makebox[\columnwidth]{ }}

\ifFGfinal
\thispagestyle{empty}
\pagestyle{empty}
\else
\author{Anonymous FG2021 submission\\ Paper ID \FGPaperID \\}
\pagestyle{plain}
\fi
\maketitle

\begin{abstract}
The majority of adversarial attack techniques perform well against deep face recognition when the full knowledge of the system is revealed (\emph{white-box}). However, such techniques act unsuccessfully in the gray-box setting where the face templates are unknown to the attackers. In this work, we propose a similarity-based gray-box adversarial attack (SGADV) technique with a newly developed objective function. SGADV utilizes the dissimilarity score to produce the optimized adversarial example, i.e., similarity-based adversarial attack. This technique applies to both white-box and gray-box attacks against authentication systems that determine genuine or imposter users using the dissimilarity score. To validate the effectiveness of SGADV, we conduct extensive experiments on face datasets of LFW, CelebA, and CelebA-HQ against deep face recognition models of FaceNet and InsightFace in both white-box and gray-box settings. The results suggest that the proposed method significantly outperforms the existing adversarial attack techniques in the gray-box setting. We hence summarize that the similarity-base approaches to develop the adversarial example could satisfactorily cater to the gray-box attack scenarios for de-authentication.
\end{abstract}

\section{Introduction}
\label{Introduction}
Deep-learning-based approaches, e.g., the deep convolutional neural networks (DCNNs), have become a de-facto standard for face recognition (FR) due to their superior accuracy. However, recent studies reveal that DCNNs are vulnerable to adversarial attacks. More specifically, the adversarial attack refers to a scheme used to fool the DCNNs in decision-making by supplying deceptive input known as an adversarial example \cite{Kur16}. In principle, the adversarial attack exploits the gradient or statistical information to generate adversarial examples in order to expose the vulnerability of DCNN-based FR \cite{Goo18}.

Among various types of adversarial attacks in literature, our proposed technique is characterized as a targeted test-time evasion attack in either white-box or gray-box scenarios:
\begin{itemize}
    \item \textit{Targeted attack:} The goal of a targeted attack is to slightly perturb the source image such that the generated adversarial example will be classified as a target label. \cite{wang2020backdoor,Goo14,Mad17,Moo16,Car171,Car17}. It is even worse for the authentication system since the adversarial example can fool the system to gain illegal access.
	\item \textit{White-box attack and gray-box attack:} The white box attack requires full knowledge of the system, including the model, data, and even the defender against threats \cite{Goo14,Mad17,Moo16,Car171,Car17,Sze13,Kur16}. A machine learning model should be secure from white-box attacks, which is the desired property \cite{Tra17}. On the other hand, the gray box attack merely requires partial knowledge instead of full knowledge \cite{xiao2018generating,Deb19}, assuming the database is not compromised in our context. Gray-box adversarial attack is believed to be more practical for attackers. 
	\item \textit{Test-time evasion:} It aims to generate the adversarial example with slight perturbation to evade human inspection but fool the classifier \cite{Goo14,Mad17,Moo16,Car171,Car17,Sze13,Big13}.
\end{itemize}

\def\sizeIntro{0.74in}
\begin{figure}
    \centering
    \setlength{\tabcolsep}{0.5mm}
    \begin{tabular}{:c:ccc:cc:}
        Source image&&Difference&&\multicolumn{2}{c:}{Adversarial example}\\
        \includegraphics[width=\sizeIntro]{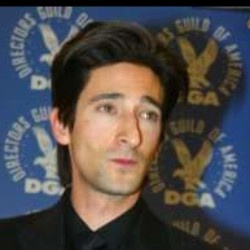}&\multirow{-5}{*}{+}&\includegraphics[width=\sizeIntro]{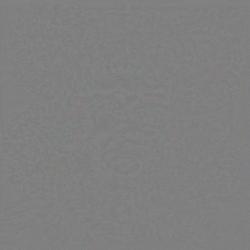}&\multirow{-5}{*}{=}&\multicolumn{2}{c:}{\includegraphics[width=\sizeIntro]{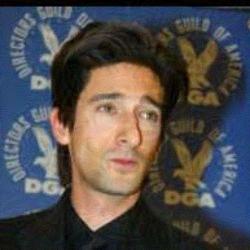}}\\
        \includegraphics[width=\sizeIntro]{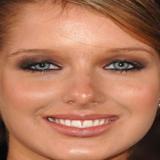}&&&&\includegraphics[width=\sizeIntro]{./figures/targeted.jpg}&\includegraphics[width=\sizeIntro]{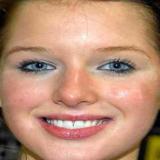}\\
        Target image&&&&White box&Gray box
    \end{tabular}
    
    \caption{Our proposed SGADV for the white-box attack and gray-box attack. The adversarial example is produced by slightly perturbing the source image based on the target image. In the white-box setting, the adversarial example is used to attack the target image and achieves 100\% attack success rate. In the gray-box setting, the adversarial example attacks another image from the same target user and achieves a 98.7\% attack success rate. The result in the gray-box setting significantly outperforms that of the label-based adversarial attacks, which gain only 17.28\% attack success rate.}
    \label{fig_example}
\end{figure}

Most existing adversarial attack techniques \cite{wang2020backdoor,Goo14,Mad17,Moo16,Car171,Car17,Sze13,Kur16,Big13} are designed to attack classification systems, whose output is the predicted label. We define these techniques as \textit{label-based} adversarial attacks as they rely on label information. However, in biometric authentication systems, e.g., FR authentication systems, no label information of the target users are stored. Hence, existing label-based adversarial attacks could not directly apply to FR-based authentication systems. Therefore, a conditional binary cross-entropy (C-BCE) objective function is proposed for the label-based adversarial attack to produce a binary output (i.e., 0/1 or false/true), resembling the binary outputs (i.e., rejection or acceptance) of biometric authentication. Although C-BCE enables binary outputs, the label-based adversarial attack shows the feasibility only when full knowledge of the system (white box) is known. Without compromising templates in the database (gray box), the attack success rate significantly decreases. The theoretical proof is given in Section \ref{renewability}.

To fill the gap in the gray-box setting, we propose the similarity-based gray-box adversarial attack (SGADV), as indicated in Fig. \ref{fig_example}. The \emph{similarity} in our context refers to the (dis)similarity score that is used to optimize the adversary to produce an adversarial example that is closest to the target image in feature space. SGADV can illegally fool the FR-based authentication in either white-box or gray-box setting, which indicates generating an adversarial example by a single target image when:
\begin{itemize}
    \item \textit{Scenario 1 (S1) - White box:} assuming the identical target image is enrolled in the database.
    \item \textit{Scenario 2 (S2) - Gray box:} assuming the same target user enrolls with any image apart from this target image.
\end{itemize}
To cater to the two attack scenarios mentioned above, a novel objective function is proposed in this paper. To further show the generalization of our proposed SGADV, we conduct experiments using three face datasets containing images of various qualities and two deep FR models. The results are compared to four benchmarking label-based adversarial attack techniques implemented with our proposed C-BCE objective function. It shows that SGADV effectively breaks FR-based authentication in both white-box and gray-box settings and significantly outperforms label-based techniques in the gray-box setting (see Table \ref{tab_compare_traditional_facenet} and \ref{tab_compare_traditional_insightface}). Our code is released at https://github.com/azrealwang/SGADV.

The contributions of this paper are summarised as follows:
\begin{itemize}
    \item To address the limitation of existing adversarial attacks against FR authentication system, we propose a novel adversarial attack technique, namely SGADV, that successfully breaks the FR-based authentication in the setting of both white box (100\% attack success rate) and gray box (up to 98.74\% attack success rate).
    \item We develop a new objective function using (dis)similarity score to increase the attack performance in the gray-box setting.
    \item We propose a C-BCE objective function to implement the benchmarking label-based adversarial attacks as the baseline against the FR-based authentication, serving as a new tool for the adversarial machine learning research community.
    \item We experimentally demonstrate that our SGADV defeats four benchmarking label-based adversarial attack techniques (minimum 78.08\% attack success rate increase). We further show the generalization of SGADV on two deep FR models and three datasets.
\end{itemize}

\section{Backgrounds}
\label{Literature}

\subsection{Adversarial attack}
\label{adversaral_attacks}
A variety of adversarial attack techniques has been proposed to attack classification systems. Some well-known methods are served as benchmarks for performance comparison, e.g., the fast gradient sign method (FGSM) \cite{Goo14}, projected gradient descent (PGD) \cite{Mad17}, DeepFool \cite{Moo16} and Carlini \& Wagner attack (CW) \cite{Car171,Car17}. The first adversarial attack on the MNIST dataset \cite{LeC98} was proposed by Biggio et al., \cite{Big13} in 2013 to mislead the classifiers, e.g., Support Vector Machine (SVM). In the same year, Szegedy et al. successfully generated adversarial examples to fool the deep-neural-networks-based classifiers \cite{Sze13}. These pioneering works light the security concerns of deep learning models. Thereafter, FGSM was proposed by Goodfellow et al. to produce adversarial examples instantly. It realizes the demands of generating a large number of adversarial examples for training and verification. In 2016, DeepFool was proposed by Moosavi-Dezfooli et al. This technique introduces slight perturbation to the image to attack deep learning models. It reveals that the deep learning models are not robust even with slight perturbation. Then, in 2017, PGD was proposed as an iterative version of the one-step attack FGSM to find the \textit{most adversarial examples} \cite{Mad17} with the acceptable efficiency. In addition, CW \cite{Car171} is deemed as one of the most aggressive adversarial attack, which was the first technique designed to counterattack the defences of \cite{Sze13} and \cite{Goo14}. It experimentally demonstrated the feasibility against ten adversarial example detection methods \cite{Car17}. Eventually, CW becomes the benchmark for evaluating the security of deep learning models and comparison with other techniques, though it is computationally exhausted.

\subsection{Adversarial attack on face recognition}
Adversarial attack techniques addressed in Section \ref{adversaral_attacks} are popularly implemented in practice. However, none of them are designed for FR. In this section, several recently proposed white-box and gray-box adversarial attacks on FR are reviewed.

The first gradient-based adversarial attack on FR was proposed by Sharif et al. \cite{sharif2016accessorize}. Sharif et al. dodged FR systems and impersonated other identities wearing an eyeglass frame with a malicious texture. It also demonstrated that the vulnerability of DCNNs could be physically exploited. Rozsa et al. proposed the first adversarial attack against deep features on FR \cite{rozsa2017lots}. It forms adversarial examples which mimic the deep features of the target image by iterative layerwise learning. Dabouei et al. proposed an efficient algorithm \cite{dabouei2019fast} directly manipulating landmarks of the face images to produce geometrically-perturbed adversarial examples. Yang et al. \cite{Yan21} proposed a GAN-based adversarial attack in the gray-box setting, namely \(A^{3}GN\). However, \(A^{3}GN\) requires at least five target images from each target user for training and inference. Deb et al. \cite{Deb19} proposed another GAN-based method, namely AdvFaces. Despite the GAN is trained in the white-box setting, AdvFaces performs the black-box attack, which cannot compete our gray-box adversarial model.

\subsection{Projected gradient descent}
PGD is regarded as a descendant of FGSM \cite{Goo14}. Unlike the one-step attack FGSM, PGD iteratively learns and seeks the most optimized adversarial examples. In this paper, PGD \cite{Mad17} is adopted but supervised by our proposed objective function (Section \ref{loss_function}) to learn the adversarial examples. Note that the proposed adversarial attack terminates at the stage of convergence, while PGD is stopped only when it reaches the pre-defined maximum steps. The adversarial attack process of PGD is given as follows. 

Let $\theta$ denote the parameters of a DCNN, $x$ be the input to the DCNN, $y$ be the label of $x$ and \(J(\theta,x,y)\) be the loss used to train the network. Then, let \(x'\) be the adversarial example of source $x$, \(y'\) be the label of \(x'\) and \(y_1\) be the target label. Since PGD is an iterative version of FGSM, PGD and FGSM share the identical adversarial attack process shown in eq (1). Thus, the \(L_\infty\)-bounded FGSM is to learn \(x'\) that 
\begin{equation}
    x'=x+\epsilon \cdot sign(\nabla_x J(\theta,x,y)),\ s.t. \ ||x'-x||_\infty<\epsilon,
\end{equation}
where satisfying \(y'\neq y\) (indiscriminate attack) or \(y'=y_1\) (targeted attack).
\(\epsilon\) is the size of the perturbation and expects to be small enough to evade human inspection. Inherent from FGSM, the \(L_\infty\)-bounded PGD is utilized by
\begin{equation}
    \begin{aligned}
    x^{t+1}=Clip_{x,\epsilon}\{x^{t}+\alpha \cdot sign(\nabla_{x^t} J(\theta,x^{t},y))\},\\
    s.t. \ ||x'-x||_\infty<\epsilon.
    \end{aligned}
    \label{eq_pgd}
\end{equation}
\(Clip(\cdot)\) denotes the function to perturb the image at each iteration step with a step size $\alpha$. The maximum steps $t_{max}$ shall satisfy \(t_{max}\geq \frac{\epsilon}{\alpha}\), to guarantee that the perturbation can reach the border. Specifically, \(x^0\) is initialized with randomly chosen starting points.

\section{Similarity-based gray-box adversarial attack}
\label{design}
This section demonstrates our proposed technique. Section \ref{adv_model} introduces the adversarial model of the proposed SGADV. Section \ref{renewability} theoretically analyzes the gap between similarity-based and label-based adversarial attacks. Section \ref{loss_function} proposes a novel objective function, and the algorithm of SGADV is presented in Section \ref{Algorithm}.

\subsection{Adversarial model}
\label{adv_model}
The FR-based authentication system generally consists of two key components, i.e., the deep FR model (e.g., DCNN) for feature embedding and the database for template storing. Our proposed SGADV applies to both white-box and gray-box scenarios:
\begin{itemize}
    \item \textit{White box:} DCNN and database are both compromised.
    \item \textit{Gray box:} The adversary is conducted in the white-box setting with respect to the DCNN but in the black-box setting with respect to the database (a face image of the target user is still required).
\end{itemize}

The white box is the baseline scenario of adversarial attacks. The systems, particularly which require high confidentiality, have to be secure in the worst case (white-box, or even strong white-box \cite{Car17}), since flaws of security are not acceptable even if an extremely small probability, which is agreed by Tramèr et al. \cite{Tra17} that security against white-box attacks is the desirable property of machine learning models. Miller et al. \cite{Mil20} also debated that a reasonable adversarial attack for evaluation should be in the white box to the system and the black box to the defense. Carlini and Wagner \cite{Car17} even asserted (without elaboration) the strong white-box requirement of a defense that the detection cannot be evaded even if it is known by the attacker. 

However, people criticize that the white-box setting is unlikely practical as it requests full knowledge of the system \cite{Deb19,dong2019efficient,zhong2020towards}. Thus, our gray-box (without knowing the database) adversarial attack can alleviate the above criticism. Specifically, pre-trained models for the well-known deep FR methods are publicly available, i.e., the true model is possibly the white box to attackers. If the true model is unavailable, a surrogate model can be learned by the black-box adversarial attack techniques  \cite{Mil20}. When either the true or surrogate model is available, our gray-box adversarial attack is viable since it merely requires the knowledge of the DCNN.

\begin{figure}[t!]
	\centering
		\includegraphics[width=3.4in]{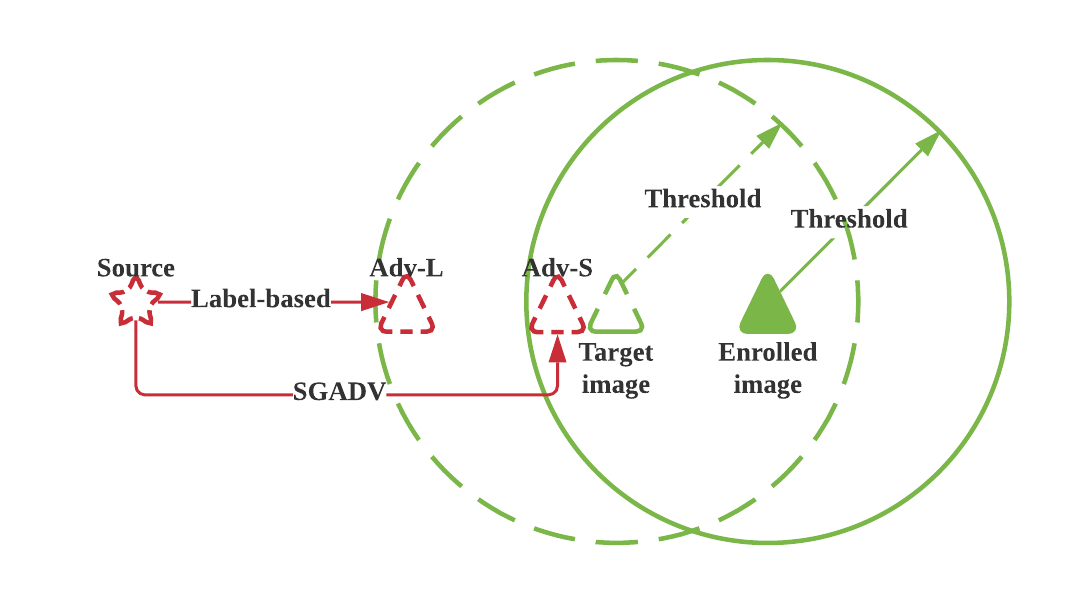}
		\caption{Comparison between label-based adversarial attack and proposed SGADV dealing with the gray-box setting. "Adv-L" and "Adv-S" represent the adversarial examples from the same target image by the label-based technique and our proposed similarity-based SGADV, respectively. The circle refers to the space where the face images are identified as the same user. The target image is from the same user with the enrolled image. As observed, both "Adv-L" and "Adv-S" are successfully categorized as the target user with the target image. However, only "Adv-S" is viable to attack the enrolled image.}
		\label{fig_One_gap_of_existing_AL_attacks_and_the_solution_proposed_in_this_project}
\end{figure}

\subsection{Challenge of adversarial attack on deep FR}
\label{renewability}
Although the FR-based authentication is converted to a binary classification task to implement the label-based adversarial attack, the label-based adversarial attack performs poorly in the gray-box setting. In contrast, our proposed similarity-based SGADV is feasible for both white-box and gray-box settings, which is theoretically analyzed as follows.

Let \(X\) be the source face image of the adversary, which will be perturbed to produce the adversarial example \(X^{adv}\). Let \(X_1^1\) represent the enrolled image of an authorized user \(X_1\) (target user). Thus, \(X_1^1\) is stored in the database and unknown to the adversary in the gray-box setting. Let \(X_1^2,X_1^3,\ldots\) be other face images of the same identity (target images), which are easily collected in the real world, e.g., from the Internet. Let \(f\left(\cdot \right)\) denote the function of feature embedding for deep FR, whose outputs are feature vectors (face templates) for comparison. Let \(||\cdot||\in[0,1]\) donate the normalized distance, which equals the dissimilarity ranged from 0 to 1 (lower indicates more similarity). Thus, in FR-based authentication systems, a query face \(X_1^2\) is recognized as \(X_1\) when
\begin{equation}
    ||f(X_1^2)-f(X_1^1)||\leq\tau,
    \label{eq_fr}
\end{equation}
where $\tau$ is a pre-defined dissimilarity threshold.

Fig. \ref{fig_One_gap_of_existing_AL_attacks_and_the_solution_proposed_in_this_project} illustrates the challenge of using the label-based adversarial attack in the gray-box setting, while our proposed SGADV is significantly enhanced. To differentiate from the proposed SGADV, the adversarial example produced by the label-based adversarial attack is represented by \(X^{adv}_L\). It is regarded as a successful attack (i.e., misclassification) when the dissimilarity between the adversarial example \(X^{adv}_L\) and target image \(X_1^2\) in feature space is less than the threshold. The adversary then can be aborted as shown in Fig. \ref{fig_One_gap_of_existing_AL_attacks_and_the_solution_proposed_in_this_project} of "Adv-L", which satisfies
\begin{equation}
    ||f(X^{adv}_L)-f(X_1^2)||\leq\tau.
    \label{eq_misclassification}
\end{equation}
However, the dissimilarity may dramatically increase when the adversarial example \(X^{adv}_L\) is compared to the enrolled image \(X_1^1\). Formulating with \eqref{eq_fr} and \eqref{eq_misclassification}, we get
\begin{equation}
    ||f(X^{adv}_L)-f(X_1^1)||\leq 2\tau.
    \label{eq_invalid}
\end{equation}
As aforementioned, the successful attack requires the dissimilarity to be less than the threshold. Thus, computed from \eqref{eq_invalid}, the probability of success of label-based adversarial attack in the gray-box setting $p_L^{success}$ should be
\begin{equation}
    p_L^{success}\approx\frac{\tau}{2\tau}=\frac{1}{2},
    \label{eq_prob_label}
\end{equation}
assuming \(||f(X^{adv}_L)-f(X_1^1)||\) follows uniform distribution \(||f(X^{adv}_L)-f(X_1^1)||\sim U(0,2\tau)\).

In contrast, the proposed SGADV elaborates dissimilarity in the objective function to find the closest adversarial example (as shown in Fig. \ref{fig_One_gap_of_existing_AL_attacks_and_the_solution_proposed_in_this_project} of "Adv-S"). The adversarial example \(X^{adv}\) satisfies
\begin{equation}
    ||f(X^{adv})-f(X_1^2)||=\ell,
    \label{eq_misfr}
\end{equation}
where $\ell$ indicates a small error when the adversary is terminated and \(\ell \ll \tau\). Thus, when comparing this adversarial example \(X^{adv}\) to the enrolled image \(X_1^1\), the threat of \(X^{adv}\) preserves in a large probability as shown in Fig. \ref{fig_One_gap_of_existing_AL_attacks_and_the_solution_proposed_in_this_project}, as
\begin{equation}
    ||f(X^{adv})-f(X_1^1)||\leq \ell+\tau.
    \label{eq_valid}
\end{equation}
In this case, the probability of success of the similarity-based adversarial attack in the gray-box setting $p^{success}$ is formulated as
\begin{equation}
    p^{success}\approx\frac{\tau}{\ell+\tau}\gg p_L^{success}\approx\frac{1}{2},
    \label{eq_prob_similarity}
\end{equation}
assuming \(||f(X^{adv})-f(X_1^1)||\) follows uniform distribution \(||f(X^{adv})-f(X_1^1)||\sim U(0,\ell+\tau)\).

Interpreted from \eqref{eq_prob_similarity}, the proposed similarity-based SGADV is more likely to break the FR-based authentication than the label-based adversarial attack in the gray-box setting. However, the probability $p^{success}$ is still slightly lower than 100\%. Thus, the attack performance in the gray-box setting is expected to degrade, comparing to the white box (experimentally demonstrated in Section \ref{performance}).

\subsection{Objective function}
\label{loss_function}
The objective function of SGADV is formulated using dissimilarity. Specifically, define \(X^{adv}\) as \(X\) with perturbation \(\delta\)
\begin{equation}
    X^{adv}=X+\delta,\ s.t.\ ||\delta||_\infty<\epsilon,
\end{equation}
where $\epsilon$ is the pre-defined size of the perturbation and small enough to evade human inspection. In FR-based authentication systems, \(X^{adv}\) is recognized as \(X_{1}\) when
\begin{equation}
    ||f(X^{adv})-f(X_1)||\leq\tau,
\end{equation}
where $\tau$ is a pre-defined dissimilarity threshold. Thus, the proposed similarity-based attack aims to find
\begin{equation}
    {\arg\min}_{X^{adv}}||f(X^{adv})-f(X_1)||.
\end{equation}
To achieve this goal, the objective function of SGADV is formulated as
\begin{equation}
    J_{SG}(X^{adv},X_1)=||f(X^{adv})-f(X_1)||.
    \label{eq_loss_st}
\end{equation}

\begin{algorithm}[!t]
    \DontPrintSemicolon
    \caption{SGADV}
    \label{alg_st}
    \KwIn{\textit{Target:} Face image $X_1$, DCNN \(f(\cdot)\); \textit{Source:} Face image $X$; \textit{Settings:} size of the perturbation $\epsilon$, step size $\alpha$, maximum steps $t_{max}$, convergence threshold $\tau_{conv}$}
    \KwOut{Adversarial example $X^{adv}$}
    \BlankLine
    \Begin(Initialization){\label{alg_Initialization_start}
        \(\delta^0\sim U(-\epsilon,\epsilon)\)\\
        \(X^0:=X+\delta^0\)\label{alg_Initialization_end}
        }
    \Repeat{Satisfy \eqref{eq_convergence_c1} or  \eqref{eq_convergence_c2} or \(t=t_{max}\)}{\label{alg_loop_start}
        \(J_{SG}(X^{t},X_1)=||f(X^{t})-f(X_1)||\)\label{alg_line_loss}\\
        \(X^{t+1}=Clip_{X,\epsilon}\{X^{t}+\alpha \cdot sign(\nabla_{X^{t}}J_{SG})\}\)\label{alg_line_pgd}
        }\label{alg_loop_end}
    \Return{\(X^{adv}:=X^{t_{stop}}\)}\label{alg_line_return}
\end{algorithm}

\begin{table*}[!t]
    \caption{Settings of deep FR and datasets}
    \label{tab_dfr_dataset}
    \centering
    \setlength{\tabcolsep}{0.5mm}{\begin{tabular}{|c|c|c|c|c|c|c|c|c|c|c|}
        \hline
        \multirow{2}{*}{Deep FR}&\multirow{2}{*}{DCNN}&\multirow{2}{*}{Loss function}&\multicolumn{2}{c|}{Pre-trained data}&\multicolumn{4}{c|}{System Setup}&\multicolumn{2}{c|}{Source}\\
        \Xcline{4-11}{0.4pt}
        &&&Dataset&Image size&Dataset&Image size&Threshold $\tau$&EER(\%)&Dataset&Image size\\
        \hline
        \hline
        FaceNet \cite{Sch15}&Inception ResNet v1&Triplet loss&VGGFace2&\(160\times160\)&\multirow{2}{*}{CelebA \cite{Liu15}}&\(160\times160\)&0.2967&1.20&\multirow{2}{*}{LFW \cite{Hua08}}&\multirow{2}{*}{\(250\times250\)}\\
        \Xcline{1-5}{0.4pt}
        \Xcline{7-9}{0.4pt}
        InsightFace \cite{Den19}&LResNet100&ArcFace loss&MS1M-Arcface&\(112\times112\)&&\(112\times112\)&0.4146&6.23&&\\
        \hline
    \end{tabular}}
\end{table*}

\subsection{Algorithm of SGADV}
\label{Algorithm}
The algorithm of SGADV is derived from \eqref{eq_pgd} of PGD \cite{Mad17}, supervised by our proposed objective function in Section \ref{loss_function}, illustrated in Algorithm \ref{alg_st}. It consists of a random initialization (Step \ref{alg_Initialization_start}-\ref{alg_Initialization_end}), and an iterative optimization (Step \ref{alg_loop_start}-\ref{alg_loop_end}) by \eqref{eq_loss_st} and \eqref{eq_pgd}. The optimization will be aborted when it satisfies any of three stop criteria (Step \ref{alg_loop_end}), which are
\begin{itemize}
    \item achieving maximum steps, i.e., \(t=t_{max}\).
    \item achieving convergence, i.e., the change of loss value is slight. To evaluate the convergence or divergence, the variation of loss value $\Delta$ between continuous steps are first recorded as
    \begin{equation}
        \begin{aligned}
            &\Delta J^{t+1}=J(X^t,*)-J(X^{t+1},*),\\
            &S=\{\Delta J^{t},\Delta J^{t+1},\Delta J^{t+2},\Delta J^{t+3},\Delta J^{t+4}\},
        \end{aligned}
        \label{eq_convergence_delta}
    \end{equation}
    where $S$ is a set of $\Delta$ with latest five steps. Then, this stop criterion is defined as 
    \begin{equation}
        \forall\Delta\in S,|\Delta|\leq\tau_{conv},
        \label{eq_convergence_c1}
    \end{equation}
    where $\tau_{conv}$ is the pre-defined convergence threshold.
    \item achieving settlement within an error range around a value. In this case, the loss value increases and decreases regularly. Thus, the stop criterion is defined as
    \begin{equation}
        \exists\Delta^*,\Delta^{**}\in S, \Delta^*\leq 0 \wedge \Delta^{**}\leq 0
        \label{eq_convergence_c2}
    \end{equation}
\end{itemize}
Ultimately, the perturbed example at the stage satisfying stop criteria is output as an adversarial example $X^{adv}$ (Step \ref{alg_line_return}).

\section{Experiments}
\label{Experiments}
\subsection{Experimental settings}
\subsubsection{Deep face recognition models}
To evaluate the generalization of our SGADV on distinct DCNNs, we conduct experiments using two popular deep FR models, i.e., FaceNet \cite{Sch15} and InsightFace \cite{Den19}. Face
images are converted to 512-dimensional templates after
feature embedding. The settings of the two models are listed in Table \ref{tab_dfr_dataset}.

\subsubsection{Datasets}
To evaluate the generalization of SGADV on various data, e.g., attacking target users of dataset A using source images of dataset B, three face datasets, i.e., Labeled Faces in the Wild (LFW) \cite{Hua08}, CelebA \cite{Liu15} and CelebA-HQ \cite{Kar17}, are adopted for experiments. Apart from the study on image size, the subjects in CelebA are selected as target users and the images in LFW are perturbed as source images. The settings are listed in Table \ref{tab_dfr_dataset}. Particularly for the study on the impact of image size, a high-quality version of CelebA dataset (size of $1024\times1024$), namely CelebA-HQ \cite{Kar17}, is adopted as the dataset of source images. In contrast, the subjects in the LFW dataset are selected as target users. The source images are then resized to various qualities from $1000\times1000$ to $112\times112$.  The experimental results on various data regarding the generalization of proposed SGADV are presented in Section \ref{performance}.

\subsubsection{Benchmarking label-based adversarial attacks}
To evaluate the attack performance, the proposed SGADV is compared against several label-based adversarial attack techniques, as shown in Table \ref{tab_attack}. The details of the comparison are presented in Section \ref{performance}. The cross-entropy objective function used in original label-based techniques does not apply to FR due to the lack of label information. Instead, we propose the C-BCE objective function for those label-based techniques to replace the cross-entropy objective function. To make the function applicable, FR-based authentication is converted to a task of binary classification as
\begin{equation}
    y=
    \begin{cases}
        0, &rejection\\
        1, &acceptance 
    \end{cases}.
\end{equation}
The label of the adversarial example $X^{adv}$ is defined as
\begin{equation}
    y^{adv}=
    \begin{cases}
        0, &||f(X^{adv})-f(X_1)||>\tau\\
        1, &||f(X^{adv})-f(X_1)||\leq\tau
    \end{cases},
    \label{eq_adv_y}
\end{equation}
where $\tau$ is the threshold. Then, the binary cross-entropy objective function is defined as
\begin{equation}
    H(p,y)=-y\log p-(1-y)\log(1-p),
    \label{eq_bce}
\end{equation}
where $y\in\{0,1\}$ represents the expected label and $p$ represents the probability of \(y=1\). Specifically in a targeted adversarial attack, \(y\equiv 1\). Meanwhile, based on  \eqref{eq_adv_y}, the probability of \(y^{adv}=1\) is computed as
\begin{equation}
    \begin{aligned}
        &p_{y^{adv}=1}=\\
        &\begin{cases}
            \frac{1-||f(X^{adv})-f(X_1)||}{1-\tau},&||f(X^{adv})-f(X_1)||>\tau\\
            1,&||f(X^{adv})-f(X_1)||\leq\tau
        \end{cases}.
    \end{aligned}
    \label{eq_adv_p}
\end{equation}
Finally, the proposed C-BCE objective function is reformulated by combining \eqref{eq_adv_y}, \eqref{eq_bce} and \eqref{eq_adv_p} as
\begin{equation}
    \begin{aligned}
        &J_{C-BCE}(X^{adv},X_1)=\\
        &-\log\frac{\min\{(1-||f(X^{adv})-f(X_1)||),(1-\tau)\}}{1-\tau}.
    \end{aligned}
    \label{eq_loss_cbce}
\end{equation}

\begin{table}[!t]
    \caption{Settings of adversarial attacks}
    \label{tab_attack}
    \centering
    \setlength{\tabcolsep}{1.3mm}{\begin{tabular}{|c|c|}
        \hline
        Technique&Settings\\
        \hline
        \hline
        FGSM \cite{Goo14}&$\epsilon=0.03$\\
        \hline
        DeepFool \cite{Moo16}&$\epsilon=0.03$, $t_{max}=40$, overshoot $\eta=0.02$\\
        \hline
        CW \cite{Car171}&\begin{tabular}{c}$\epsilon=0.03$, $\alpha=0.001$, $t_{max}=1000$,\\binary search iterations = 20\end{tabular}\\
        \hline
        PGD \cite{Mad17}&$\epsilon=0.03$, $\alpha=0.001$, $t_{max}=40$\\
        \Xcline{1-2}{0.8pt}
        SGADV&$\epsilon=0.03$, $\alpha=0.001$, $t_{max}=1000$, $\tau_{conv}=0.0001$\\
        \hline
    \end{tabular}}
\end{table}

\subsubsection{Evaluation metrics}
\begin{itemize}
	\item \textit{Equal error rate (EER)} and \textit{threshold} are computed as the system setup. The attack success rate is expected largely greater than EER, and dissimilarity is no more than the threshold.
	\item \textit{False positive rate (FPR)} and \textit{true positive rate (TPR)} are collected to obtain ROC curves illustrating the security under the adversarial attack.
	\item \textit{Attack success rate (ASR)} refers to the ratio of adversarial examples that successfully access the system. For an adversarial example from a target user $X_1$ with $N$ face images, ASR is computed as
	\begin{equation}
	    ASR = \frac{1}{N}\Sigma_{i=1}^{N}(||f(X^{adv}-f(X_1^i)||\leq\tau).
	\end{equation}
	ASR is the most critical index of security and attack performance. The higher ASR indicates better attack performance and poorer security of the system.
	\item \textit{Dissimilarity} indicates the closeness between the adversarial example and \emph{target image} in feature space. It exhibits the success of the attack when the dissimilarity is less than the threshold. In this paper, \(dissimilarity\in[0,1]\) is computed by the normalized cosine distance,
	\begin{equation}
	    dissimilarity=0.5\times cos[f(X^{adv}),f(X_1)]+0.5.
	\end{equation}
	\item \textit{Structural similarity index measure (SSIM)} \cite{wang2004image} is a quantitative perceptual metric to evaluate the impact of perturbation (i.e., the image quality). A higher value indicates the adversarial example is slightly perturbed to evade human inspection better.
	\item \textit{Learned perceptual image patch similarity (LPIPS)} \cite{zhang2018perceptual} is another perceptual metric to measure the distance between image patches. The higher value shows the more difference between patches.
	\end{itemize}

\subsubsection{Attack scenarios}
\label{scenarios}
Settings of white-box and gray-box attack scenarios are detailed as followed. The definition of each scenario refers to Section \ref{Introduction}.
\begin{itemize}
    \item \textit{S1 - white box:} There are 1580 adversarial examples (source and target image pairs) in total for ASR computing, including 158 subjects with ten face images each selected from LFW and CelebA datasets.
    \item \textit{S2 - gray box:} When one among ten images is selected as the target image (not enrolled in the system), the adversarial example learned from this target image is to attack the rest nine images (possibly enrolled in the system). Then, ten-fold cross-validation for ten images of each subject is conducted. In total, there are 1580 adversarial examples produced, and each adversarial example attacks nine samples.
\end{itemize}

\subsubsection{Implementation}
The implementation of the experiments is derived from a project, namely foolbox \cite{rauber2017foolboxnative,rauber2017foolbox}. The machine used for simulation is equipped with i7-9700 CPU @ 3.00GHz, 3200 MHz 64GB RAM, NVIDIA TITAN Xp GPU @ 12GB RAM.

\begin{table*}[!t]
    \centering
        \caption{Attack performance of SGADV and label-based adversarial attacks against FaceNet \cite{Sch15}}
        \label{tab_compare_traditional_facenet}
        
        \setlength{\tabcolsep}{3.8mm}{\begin{tabular}{|c|c|c|c|c|c|c|c|c|}
            \hline
            \multirow{2}{*}{Technique}&\multirow{2}{*}{Threshold $\tau$}&\multirow{2}{*}{EER(\%)}&\multicolumn{2}{c|}{ASR(\%)}&\multirow{2}{*}{Dissimilarity}&\multirow{2}{*}{SSIM}&\multirow{2}{*}{LPIPS}&Time cost(s)\\
            \Xcline{4-5}{0.4pt}
            &&&White box&Gray box&&&&per example\\
            \hline
            \hline
            FGSM \cite{Goo14}&\multirow{5.5}{*}{0.2967}&\multirow{5.5}{*}{1.20}&53.04&22.71&0.2922&0.7678&0.2059&\textbf{0.10}\\
            \Xcline{1-1}{0.4pt}
            \Xcline{4-9}{0.4pt}
            DeepFool \cite{Moo16}&&&100.00&11.72&0.2942&0.9828&0.0130&0.32\\
            \Xcline{1-1}{0.4pt}
            \Xcline{4-9}{0.4pt}
            CW \cite{Car171}&&&100.00&10.72&0.2965&\textbf{0.9986}&\textbf{0.0014}&194.42 \\
            \Xcline{1-1}{0.4pt}
            \Xcline{4-9}{0.4pt}
            PGD \cite{Mad17}&&&100.00&17.28&0.2834&0.8669&0.0738&0.72\\
            \Xcline{1-1}{0.8pt}
            \Xcline{4-9}{0.8pt}
            SGADV&&&100.00&\textbf{98.74}&\textbf{0.0032}&0.8551&0.0984&2.86\\
            \hline
        \end{tabular}}
\end{table*}

\begin{table*}[!t]
    \centering
        \caption{Attack performance of SGADV and label-based adversarial attacks against InsightFace \cite{Den19}}
        \label{tab_compare_traditional_insightface}
        
        \setlength{\tabcolsep}{3.8mm}{\begin{tabular}{|c|c|c|c|c|c|c|c|c|}
            \hline
            \multirow{2}{*}{Technique}&\multirow{2}{*}{Threshold $\tau$}&\multirow{2}{*}{EER(\%)}&\multicolumn{2}{c|}{ASR(\%)}&\multirow{2}{*}{Dissimilarity}&\multirow{2}{*}{SSIM}&\multirow{2}{*}{LPIPS}&Time cost(s)\\
            \Xcline{4-5}{0.4pt}
            &&&White box&Gray box&&&&per example\\
            \hline
            \hline
            FGSM \cite{Goo14}&\multirow{5.5}{*}{0.4146}&\multirow{5.5}{*}{6.23}&73.10&29.11&0.3844&0.8236&0.0849&\textbf{0.10}\\
            \Xcline{1-1}{0.4pt}
            \Xcline{4-9}{0.4pt}
            DeepFool \cite{Moo16}&&&100.00&11.19&0.4128&0.9953&0.0007&0.21\\
            \Xcline{1-1}{0.4pt}
            \Xcline{4-9}{0.4pt}
            CW \cite{Car171}&&&100.00&10.58&0.4144&\textbf{0.9997}&\textbf{0.0001}&87.67\\
            \Xcline{1-1}{0.4pt}
            \Xcline{4-9}{0.4pt}
            PGD \cite{Mad17}&&&100.00&15.15&0.4056&0.9157&0.0130&0.52\\
            \Xcline{1-1}{0.8pt}
            \Xcline{4-9}{0.8pt}
            SGADV&&&100.00&\textbf{93.23}&\textbf{0.0665}&0.8746&0.0458&4.23\\
            \hline
        \end{tabular}}
\end{table*}

\subsection{Performance of SGADV}
\label{performance}
As shown in Table \ref{tab_compare_traditional_facenet} and \ref{tab_compare_traditional_insightface}, the proposed SGADV successfully gains illegal access to the FR-based authentication with more than 93.23\% ASR in both white-box and gray-box settings. SGADV also achieves competitive efficiency with a maximum of 4.23s GPU time per adversarial example. It is not surprising that the adversarial examples generated by SGADV achieve 100\% ASR in the white-box setting since the dissimilarity between the adversarial examples and target images is much lower than the threshold using our proposed objective function (see the column of dissimilarity). Thus, the adversarial examples easily gain access to the system. Furthermore, as expected, the ASR of the gray-box scenarios slightly decreases compared to that of the white-box scenarios. However, it remains 98.74\% against FaceNet and 93.23\% against Insightface. It concludes that the gray-box adversarial attack is viable. The deterioration of ASR in the gray-box setting has been explained in \eqref{eq_prob_similarity} that the probability of success is slightly less than 100\%. 

\begin{figure}[!t]
    \centering
    \includegraphics[width=3.4in]{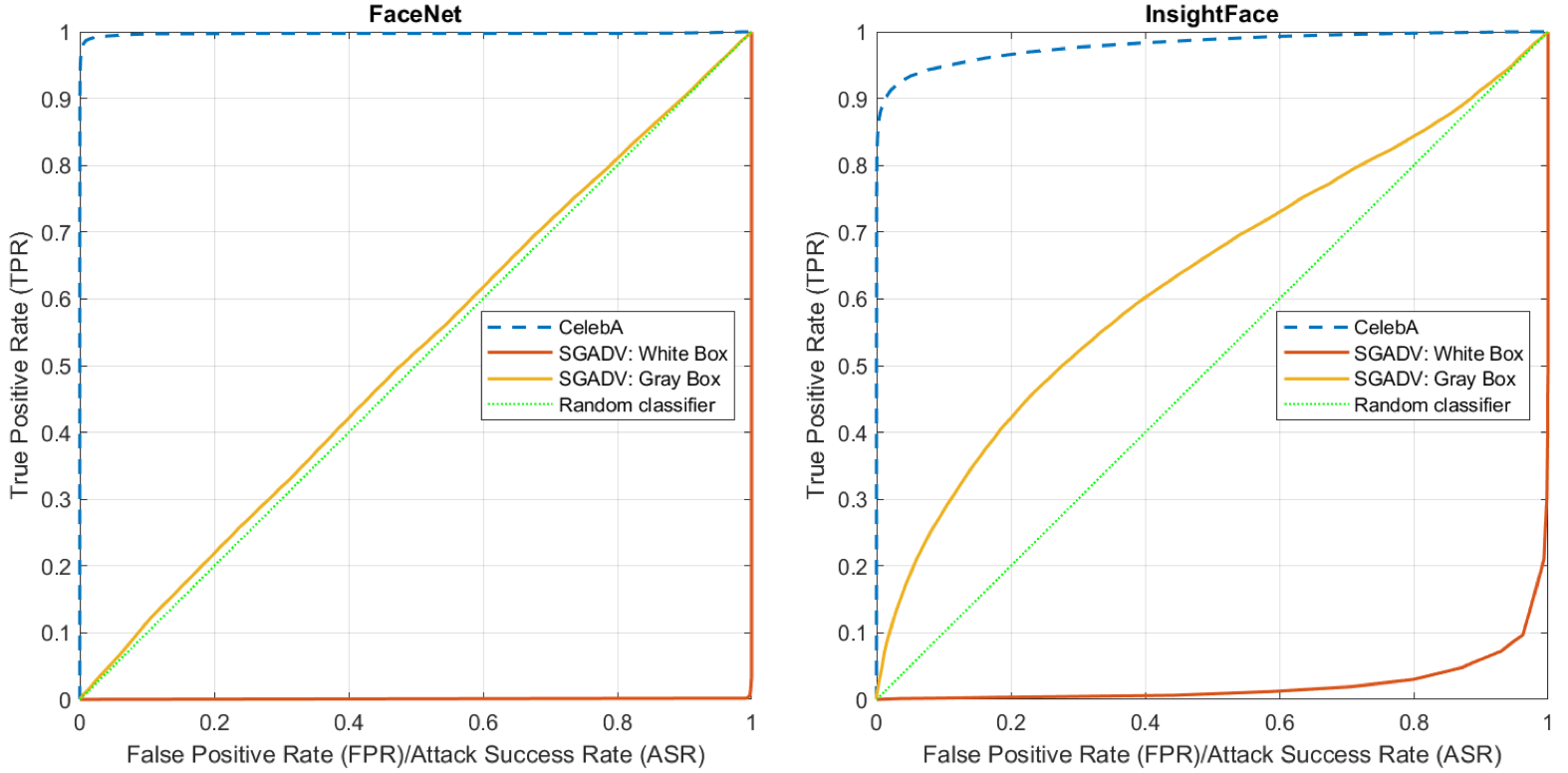}
    \caption{ROC curves of the proposed SGADV, comparing to that of the benign CelebA dataset \cite{Liu15}.}
    \label{fig_ablation_roc}
\end{figure}

\def\imagewidthDiff{0.55in}
\begin{figure}[!t]
    \centering
    \setlength{\tabcolsep}{0.5mm}
    \begin{tabular}{cccccc}
        $\epsilon$&0.003&0.01&0.03&0.1&0.3\\
        &&&&&\\
        \multirow{-3}{*}{Adv}&\includegraphics[width=\imagewidthDiff]{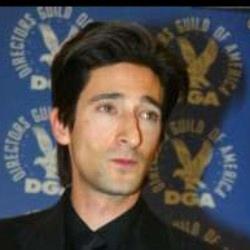}&\includegraphics[width=\imagewidthDiff]{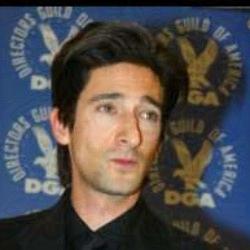}&\includegraphics[width=\imagewidthDiff]{./figures/ST_adv.jpg}&\includegraphics[width=\imagewidthDiff]{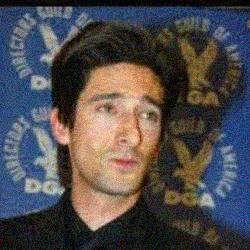}&\includegraphics[width=\imagewidthDiff]{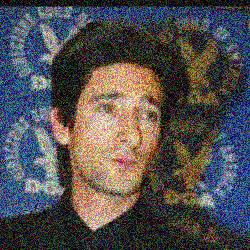}\\
        \multirow{-3}{*}{Diff}&\includegraphics[width=\imagewidthDiff]{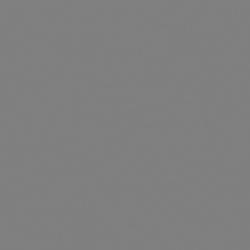}&\includegraphics[width=\imagewidthDiff]{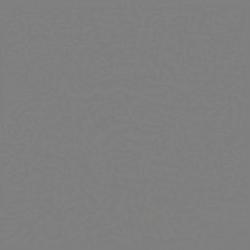}&\includegraphics[width=\imagewidthDiff]{./figures/ST_noise.jpg}&\includegraphics[width=\imagewidthDiff]{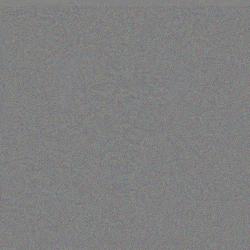}&\includegraphics[width=\imagewidthDiff]{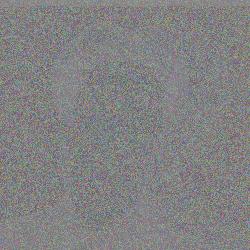}\\
        &&&&&\\
        SSIM&0.9966&0.9669&0.8551&0.4943&0.1807\\
        LPIPS&0.0013&0.0227&0.0984&0.3617&0.8791
    \end{tabular}
    \caption{Illustration of adversarial examples (Adv) and difference (Diff) in various sizes of perturbation $\epsilon$. As observed, the difference among the adversarial examples is visually negligible when \(\epsilon\leq0.03\), while the perturbation is clearly visible when \(\epsilon>0.03\). Our setting is \(\epsilon=0.03\).}
    \label{fig_ablation_epsilons_sample}
\end{figure}

\begin{table}[!t]
    \centering
        \caption{Attack performance using the source image of various sizes}
        \label{tab_ablation_imagesize}
        \setlength{\tabcolsep}{3.3mm}{\begin{tabular}{|c|c|c|c|}
            \hline
            \multirow{2}{*}{Image size}&Time cost(s)&\multicolumn{2}{c|}{ASR(\%)}\\
            \Xcline{3-4}{0.4pt}
            &per example&White box&Gray box\\
            \hline
            \hline
            $1000\times1000$&8.00&100.00&98.58\\
            \hline
            $500\times500$&\textbf{2.85}&100.00&98.62\\
            \hline
            \begin{tabular}{c}\(250\times250\)\\(Our setting)\end{tabular}&2.89&100.00&\textbf{98.70}\\
            \hline
            $160\times160$&5.48&100.00&98.21\\
            \hline
            $112\times112$&6.47&100.00&98.10\\
            \hline
        \end{tabular}}
\end{table}

The feasibility of SGADV is also illustrated by the ROC curves as shown in Fig. \ref{fig_ablation_roc}. It displays the ROC curves of the benign system without an attack (blue dash line) and the system under the attack of SGADV (red and yellow lines). Specifically, the authentication entirely fails to function under the white-box SGADV attack. It is observed that the ROC curve of the white-box SGADV (red line) lies below the random classifier (green dot line), which shows the system performance under attack is worse than randomly guessing. Under the gray-box SGADV, although ASR can be slightly degraded, the ROC curve (yellow line) shows only slightly better than random classification. It further implies that the pre-trained models are not safe for authentication.

In addition, comparing the results of SGADV in Table \ref{tab_compare_traditional_facenet} with Table \ref{tab_compare_traditional_insightface}, the attack performance of SGADV is highly correlated with the EER of the deep FR model. It is observed that ASR+EER of SGADV achieves 99.94\% for FaceNet and 99.46\% for InsightFace. Such results are caused by the extreme closeness between the adversarial examples and the target images. Consequently, the error of the benign data (i.e., EER) remains in adversarial examples. 

Moreover, to evaluate the image quality of the adversarial examples (i.e., the difference between the adversarial example and source image), SSIM and LPIPS are recorded in Table \ref{tab_compare_traditional_facenet} and \ref{tab_compare_traditional_insightface}. The results show that with a fixed setting of the size of perturbation \(\epsilon=0.03\), SSIM for SGADV remains higher than 0.855, and LPIPS remains lower than 0.1. Such results indicate that the adversarial examples are close enough to source images to evade human inspection. Moreover, as observed in Fig. \ref{fig_ablation_epsilons_sample}, the difference among the adversarial examples is visually negligible when \(\epsilon\leq0.03\), while the perturbation is clearly visible when \(\epsilon>0.03\).

Furthermore, we conduct the comparison between label-based adversarial attacks and our proposed SGADV. As  shown in Table \ref{tab_compare_traditional_facenet} and \ref{tab_compare_traditional_insightface}, SGADV significantly outperforms label-based adversarial attacks on FR in the gray-box setting regardless of the specific deep FR models. It is not difficult for the label-based adversarial attack to generate the adversarial example defeating the authentication with 100\% ASR in the white-box setting. This also experimentally demonstrates the effectiveness of the proposed C-BCE objective function on the binary classification task. However, the feasibility of label-based adversarial attacks significantly reduces in the gray-box setting. The reason is attributed to the dissimilarity between adversarial examples and target images under the label-based adversarial attacks remains only slightly lower than the threshold (see the column of dissimilarity), which recalls the challenge addressed in Section \ref{renewability}. Besides, it is noted that the image quality (SSIM and LPIPS) of the adversarial examples produced by CW is better than that by other techniques, including SGADV, though CW is computationally exhausted. However, the improvement of the image quality will be unnoticeable since the perturbation of SGADV is subtle enough to evade human inspection, as exampled in Fig. \ref{fig_example}. In addition, though SGADV is less efficient than FGSM, DeepFool, and PGD, the time cost of SGADV is still satisfied to the attacker, yet gains a significant increase on the attack performance.

In addition, we review two existing adversarial attacks against deep FR, i.e., \(A^3GN\) \cite{Yan21} and AdvFaces \cite{Deb19}. \(A^3GN\) achieves a 99.66\% ASR in the white-box setting and 98.96\% in the gray-box setting \cite{Yan21}. In contrast, the ASR of the proposed SGADV achieves 100\% in the white-box setting and 98.74\% in the gray-box setting (as shown in Table \ref{tab_compare_traditional_facenet}). We further stress that both ASRs are comparable, though the datasets used in \cite{Yan21} and SGADV are not the same. For AdvFaces, one of the attack scenarios is that \cite{Deb19} trains the GAN using FaceNet, then attacks FaceNet. In this scenario, the experimental result shows the ASR is 20.85\%, which is far behind the proposed SGADV (98.74\% against FaceNet). Nevertheless, the fairness of the comparison is a concern as AdvFaces is a black-box attack after the GAN is trained, which indicates the discrepancy between the adversarial models.

Lastly, we study the impact of input image size on efficiency and attack performance. The CelebA-HQ dataset \cite{Kar17} (original size of \(1024\times1024\)) is adopted as the dataset of source images. Then, the source images are reshaped to \(1000\times1000\), \(500\times500\), \(250\times250\), \(160\times160\), \(112\times112\), respectively, whilst images in the LFW dataset as target are aligned to \(160\times160\) for FaceNet using MTCNN \cite{zhang2016joint}. 
Surprisingly, as the results shown in Table \ref{tab_ablation_imagesize}, the source image with a slightly larger size than that of the target image outperforms other settings in terms of both efficiency and ASR. For instance, to attack a target image with the size of \(160\times160\), the source image with the size of \(250\times250\) gains the highest ASR, while with the size of \(500\times500\) costs the least time. It further implies that the larger image contains more pixels which can be perturbed in the adversarial attack. Therefore, to balance the attack performance and efficiency, the slightly larger size of the source images than that of target images is suggested. Note that the study on the impact of input image size is not conducted for InsightFace as \(112\times112\) input is requested by InsightFace.

\section{Conclusion}
\label{Conclusion}
In this paper, we propose a similarity-based adversarial attack against FR-based authentication using the newly developed objective function, namely SGADV. SGADV shows the feasibility and satisfactory efficiency in both white-box and gray-box settings. Particularly, SGADV dramatically outperforms four benchmarking label-based adversarial attack techniques in the same gray-box setting. SGADV further shows the generalization on two deep FR models and three face datasets. In addition, the experimental results reveal that the pre-trained models are not safe in practice even if the database is not compromised. As such, this study is critical to empower the users against privacy threats and prevent one's data from being taken without the user's consent. Last but not least, this work also caters to the new gray-box attack scenarios for de-authentication and further implicates the pitfalls and limitations against the robustness of DCNNs.

In future work, we intend to extend the similarity-based method to attack multiple target users or conduct studies on transferability. Besides, we are planning to design a defense to enhance the resistance against adversarial attacks.




{\small
\bibliographystyle{IEEEtran}
\bibliography{egbib}

\begin{thebibliography}{10}
\providecommand{\url}[1]{#1}
\csname url@samestyle\endcsname
\providecommand{\newblock}{\relax}
\providecommand{\bibinfo}[2]{#2}
\providecommand{\BIBentrySTDinterwordspacing}{\spaceskip=0pt\relax}
\providecommand{\BIBentryALTinterwordstretchfactor}{4}
\providecommand{\BIBentryALTinterwordspacing}{\spaceskip=\fontdimen2\font plus
\BIBentryALTinterwordstretchfactor\fontdimen3\font minus
  \fontdimen4\font\relax}
\providecommand{\BIBforeignlanguage}[2]{{%
\expandafter\ifx\csname l@#1\endcsname\relax
\typeout{** WARNING: IEEEtran.bst: No hyphenation pattern has been}%
\typeout{** loaded for the language `#1'. Using the pattern for}%
\typeout{** the default language instead.}%
\else
\language=\csname l@#1\endcsname
\fi
#2}}
\providecommand{\BIBdecl}{\relax}
\BIBdecl

\bibitem{Kur16}
A.~Kurakin, I.~Goodfellow, and S.~Bengio, ``Adversarial machine learning at
  scale,'' \emph{arXiv preprint arXiv:1611.01236}, 2016.

\bibitem{Goo18}
I.~Goodfellow, P.~McDaniel, and N.~Papernot, ``Making machine learning robust
  against adversarial inputs,'' \emph{Communications of the ACM}, vol.~61,
  no.~7, pp. 56--66, 2018.

\bibitem{wang2020backdoor}
S.~Wang, S.~Nepal, C.~Rudolph, M.~Grobler, S.~Chen, and T.~Chen, ``Backdoor
  attacks against transfer learning with pre-trained deep learning models,''
  \emph{IEEE Transactions on Services Computing}, 2020.

\bibitem{Goo14}
I.~J. Goodfellow, J.~Shlens, and C.~Szegedy, ``Explaining and harnessing
  adversarial examples,'' \emph{arXiv preprint arXiv:1412.6572}, 2014.

\bibitem{Mad17}
A.~Madry, A.~Makelov, L.~Schmidt, D.~Tsipras, and A.~Vladu, ``Towards deep
  learning models resistant to adversarial attacks,'' \emph{arXiv preprint
  arXiv:1706.06083}, 2017.

\bibitem{Moo16}
S.-M. Moosavi-Dezfooli, A.~Fawzi, and P.~Frossard, ``Deepfool: a simple and
  accurate method to fool deep neural networks,'' in \emph{Proceedings of the
  IEEE conference on computer vision and pattern recognition}, 2016, Conference
  Proceedings.

\bibitem{Car171}
N.~Carlini and D.~Wagner, ``Towards evaluating the robustness of neural
  networks,'' in \emph{2017 IEEE Symposium on Security and Privacy (SP)}, 2017,
  Conference Proceedings.

\bibitem{Car17}
------, ``Adversarial examples are not easily detected: Bypassing ten detection
  methods,'' in \emph{Proceedings of the 10th ACM Workshop on Artificial
  Intelligence and Security}, 2017, Conference Proceedings.

\bibitem{Sze13}
C.~Szegedy, W.~Zaremba, I.~Sutskever, J.~Bruna, D.~Erhan, I.~Goodfellow, and
  R.~Fergus, ``Intriguing properties of neural networks,'' \emph{arXiv preprint
  arXiv:1312.6199}, 2013.

\bibitem{Tra17}
F.~Tramèr, A.~Kurakin, N.~Papernot, I.~Goodfellow, D.~Boneh, and P.~McDaniel,
  ``Ensemble adversarial training: Attacks and defenses,'' \emph{arXiv preprint
  arXiv:1705.07204}, 2017.

\bibitem{xiao2018generating}
C.~Xiao, B.~Li, J.-Y. Zhu, W.~He, M.~Liu, and D.~Song, ``Generating adversarial
  examples with adversarial networks,'' \emph{arXiv preprint arXiv:1801.02610},
  2018.

\bibitem{Deb19}
D.~Deb, J.~Zhang, and A.~K. Jain, ``Advfaces: Adversarial face synthesis,'' in
  \emph{2020 IEEE International Joint Conference on Biometrics (IJCB)}, 2019,
  Conference Proceedings.

\bibitem{Big13}
B.~Biggio, I.~Corona, D.~Maiorca, B.~Nelson, N.~Šrndić, P.~Laskov,
  G.~Giacinto, and F.~Roli, ``Evasion attacks against machine learning at test
  time,'' in \emph{Joint European conference on machine learning and knowledge
  discovery in databases}, 2013, Conference Proceedings.

\bibitem{LeC98}
Y.~LeCun, L.~Bottou, Y.~Bengio, and P.~Haffner, ``Gradient-based learning
  applied to document recognition,'' \emph{Proceedings of the IEEE}, vol.~86,
  no.~11, pp. 2278--2324, 1998.

\bibitem{sharif2016accessorize}
M.~Sharif, S.~Bhagavatula, L.~Bauer, and M.~K. Reiter, ``Accessorize to a
  crime: Real and stealthy attacks on state-of-the-art face recognition,'' in
  \emph{Proceedings of the 2016 acm sigsac conference on computer and
  communications security}, 2016, pp. 1528--1540.

\bibitem{rozsa2017lots}
A.~Rozsa, M.~G{\"u}nther, and T.~E. Boult, ``Lots about attacking deep
  features,'' in \emph{2017 IEEE International Joint Conference on Biometrics
  (IJCB)}.\hskip 1em plus 0.5em minus 0.4em\relax IEEE, 2017, pp. 168--176.

\bibitem{dabouei2019fast}
A.~Dabouei, S.~Soleymani, J.~Dawson, and N.~Nasrabadi, ``Fast
  geometrically-perturbed adversarial faces,'' in \emph{2019 IEEE Winter
  Conference on Applications of Computer Vision (WACV)}.\hskip 1em plus 0.5em
  minus 0.4em\relax IEEE, 2019, pp. 1979--1988.

\bibitem{Yan21}
L.~Yang, Q.~Song, and Y.~Wu, ``Attacks on state-of-the-art face recognition
  using attentional adversarial attack generative network,'' \emph{Multimedia
  Tools and Applications}, vol.~80, no.~1, pp. 855--875, 2021.

\bibitem{Mil20}
D.~J. Miller, Z.~Xiang, and G.~Kesidis, ``Adversarial learning targeting deep
  neural network classification: A comprehensive review of defenses against
  attacks,'' \emph{Proceedings of the IEEE}, vol. 108, no.~3, pp. 402--433,
  2020.

\bibitem{dong2019efficient}
Y.~Dong, H.~Su, B.~Wu, Z.~Li, W.~Liu, T.~Zhang, and J.~Zhu, ``Efficient
  decision-based black-box adversarial attacks on face recognition,'' in
  \emph{Proceedings of the IEEE/CVF Conference on Computer Vision and Pattern
  Recognition}, 2019, pp. 7714--7722.

\bibitem{zhong2020towards}
Y.~Zhong and W.~Deng, ``Towards transferable adversarial attack against deep
  face recognition,'' \emph{IEEE Transactions on Information Forensics and
  Security}, vol.~16, pp. 1452--1466, 2020.

\bibitem{Sch15}
F.~Schroff, D.~Kalenichenko, and J.~Philbin, ``Facenet: A unified embedding for
  face recognition and clustering,'' in \emph{Proceedings of the IEEE
  conference on computer vision and pattern recognition}, 2015, Conference
  Proceedings.

\bibitem{Liu15}
Z.~Liu, P.~Luo, X.~Wang, and X.~Tang, ``Deep learning face attributes in the
  wild,'' in \emph{Proceedings of the IEEE international conference on computer
  vision}, 2015, Conference Proceedings.

\bibitem{Hua08}
G.~B. Huang, M.~Mattar, T.~Berg, and E.~Learned-Miller, ``Labeled faces in the
  wild: A database forstudying face recognition in unconstrained
  environments,'' in \emph{Workshop on faces in'Real-Life'Images: detection,
  alignment, and recognition}, 2008, Conference Proceedings.

\bibitem{Den19}
J.~Deng, J.~Guo, N.~Xue, and S.~Zafeiriou, ``Arcface: Additive angular margin
  loss for deep face recognition,'' in \emph{Proceedings of the IEEE/CVF
  Conference on Computer Vision and Pattern Recognition}, 2019, Conference
  Proceedings.

\bibitem{Kar17}
T.~Karras, T.~Aila, S.~Laine, and J.~Lehtinen, ``Progressive growing of gans
  for improved quality, stability, and variation,'' \emph{arXiv preprint
  arXiv:1710.10196}, 2017.

\bibitem{wang2004image}
Z.~Wang, A.~C. Bovik, H.~R. Sheikh, and E.~P. Simoncelli, ``Image quality
  assessment: from error visibility to structural similarity,'' \emph{IEEE
  transactions on image processing}, vol.~13, no.~4, pp. 600--612, 2004.

\bibitem{zhang2018perceptual}
R.~Zhang, P.~Isola, A.~A. Efros, E.~Shechtman, and O.~Wang, ``The unreasonable
  effectiveness of deep features as a perceptual metric,'' in \emph{CVPR},
  2018.

\bibitem{rauber2017foolboxnative}
\BIBentryALTinterwordspacing
J.~Rauber, R.~Zimmermann, M.~Bethge, and W.~Brendel, ``Foolbox native: Fast
  adversarial attacks to benchmark the robustness of machine learning models in
  pytorch, tensorflow, and jax,'' \emph{Journal of Open Source Software},
  vol.~5, no.~53, p. 2607, 2020. [Online]. Available:
  \url{https://doi.org/10.21105/joss.02607}
\BIBentrySTDinterwordspacing

\bibitem{rauber2017foolbox}
\BIBentryALTinterwordspacing
J.~Rauber, W.~Brendel, and M.~Bethge, ``Foolbox: A python toolbox to benchmark
  the robustness of machine learning models,'' in \emph{Reliable Machine
  Learning in the Wild Workshop, 34th International Conference on Machine
  Learning}, 2017. [Online]. Available: \url{http://arxiv.org/abs/1707.04131}
\BIBentrySTDinterwordspacing

\bibitem{zhang2016joint}
K.~Zhang, Z.~Zhang, Z.~Li, and Y.~Qiao, ``Joint face detection and alignment
  using multitask cascaded convolutional networks,'' \emph{IEEE Signal
  Processing Letters}, vol.~23, no.~10, pp. 1499--1503, 2016.

\end{thebibliography}
}

\end{document}